\documentclass[letterpaper]{article} 
\usepackage{aaai25}  
\usepackage{times}  
\usepackage{helvet}  
\usepackage{courier}  
\usepackage[hyphens]{url}  
\usepackage{graphicx} 
\usepackage{natbib}  
\usepackage{caption} 
\usepackage{amssymb}
\usepackage{bbding}
\usepackage{pifont}
\usepackage{wasysym}
\usepackage{utfsym}
\usepackage{tcolorbox}
\usepackage{algorithm}
\usepackage{algorithmic}
\usepackage{xspace}
\usepackage{newfloat}
\usepackage{listings}

\newcommand{\abbr}[0]{VATE\xspace}
\DeclareCaptionStyle{ruled}{labelfont=normalfont,labelsep=colon,strut=off} 
\floatstyle{ruled}
\newfloat{listing}{tb}{lst}{}
\floatname{listing}{Listing}
\title{AI-Driven Virtual Teacher for Enhanced Educational Efficiency: \\Leveraging Large Pretrained Models for Autonomous Error Analysis and Correction}
\author{
    Tianlong Xu$^{\rm 1}$,
    Yi-Fan Zhang$^{\rm 2}$,
    Zhendong Chu$^{\rm 1}$,
    Shen Wang$^{\rm 1}$,
    Qingsong Wen$^{\rm 1}$\thanks{Corresponding author}
}
\affiliations{
    \textsuperscript{\rm 1}Squirrel Ai Learning, \textsuperscript{\rm 2}CASIA MAIS-NLPR\\

}

\usepackage{bibentry}

\begin{document}

\maketitle

\begin{abstract}
Students frequently make mistakes while solving mathematical problems, and traditional error correction methods are both time-consuming and labor-intensive. This paper introduces an innovative \textbf{V}irtual \textbf{A}I \textbf{T}eacher system designed to autonomously analyze and correct student \textbf{E}rrors (\abbr). Leveraging advanced large language models (LLMs), the system uses student drafts as a primary source for error analysis, which enhances understanding of the student's learning process. It incorporates sophisticated prompt engineering and maintains an error pool to reduce computational overhead. The AI-driven system also features a real-time dialogue component for efficient student interaction. Our approach demonstrates significant advantages over traditional and machine learning-based error correction methods, including reduced educational costs, high scalability, and superior generalizability. The system has been deployed on the Squirrel AI learning platform for elementary mathematics education, where it achieves 78.3\% accuracy in error analysis and shows a marked improvement in student learning efficiency. Satisfaction surveys indicate a strong positive reception, highlighting the system's potential to transform educational practices.

\end{abstract}

\section{Introduction}

Students frequently make errors while solving mathematical problems. Rather than merely identifying and correcting mistakes, teachers should meticulously mark each error within a student’s answer to better understand the underlying difficulties, rather than simply marking the entire answer as incorrect. This detailed grading approach enables teachers to quickly and clearly pinpoint specific student errors, allowing for more targeted guidance. In traditional educational settings, education experts typically analyze student errors to help educators better understand these mistakes. With the advent of machine learning, some research efforts have focused on collecting large amounts of student data to identify specific error patterns~\cite{jarvis2004applying}. This data-driven approach aims to determine the most effective instructional strategies to address students’ skill deficits or misunderstandings~\cite{feldman2018automatic,haryanti2019analysis,priyani2018error}. However, existing methods face several significant limitations:

\begin{figure}
    \centering
    \includegraphics[width=\linewidth]{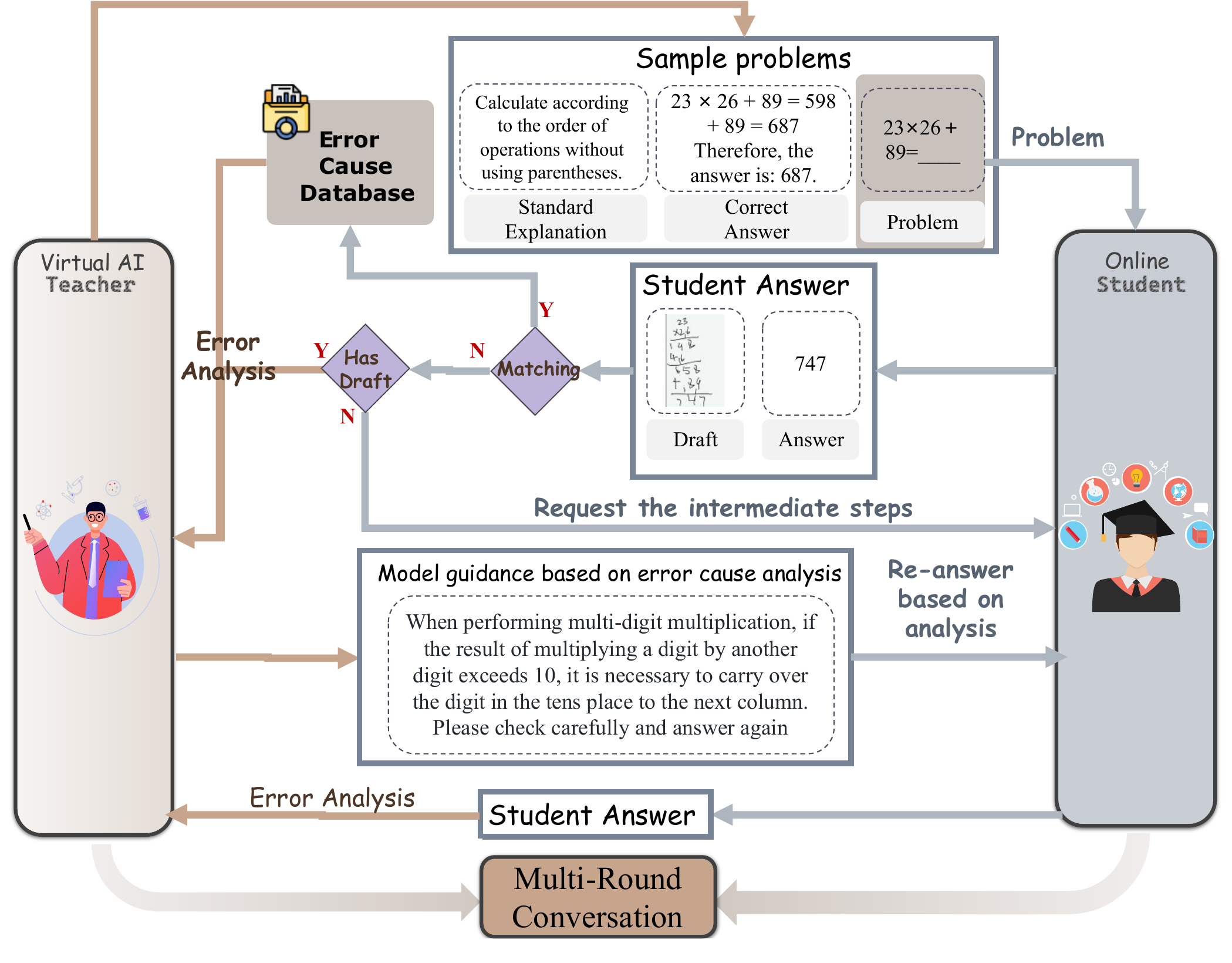}
    \caption{\textbf{The operational mode of our Virtual AI Teacher system:} After assigning problems to students, it accepts the students' drafts and answers. First, it compares them with our internal error cause database; if a matching error is found, it directly provides the error analysis. If not, it utilizes our large model system for error analysis. The system then engages in multiple rounds of dialogue with the student, guiding them to correct the corresponding errors based on the error analysis in each conversation.}
    \label{fig:function}
    \vspace{-0.1cm}
\end{figure}
\begin{enumerate}
    \item \textbf{Time-Consuming and Labor-Intensive:} They typically require education experts to provide one-on-one tutoring with students or to build systems that analyze long-term student work to identify common patterns. This process is both challenging and time-consuming~\cite{brown2016mathematics}, remaining a significant obstacle for educators.
    \item \textbf{Limited Generalizability:} Error types identified for individual students are difficult to generalize to others. Each student must undergo the same complex process, making it challenging to apply insights across a broader student population.
    \item \textbf{Restricted Applicability:} Different subjects and grade levels may require distinct systems and experts, meaning that the identified causes of errors often only apply to the specific data that has been collected.
\end{enumerate} 

Recently, large language models (LLMs), such as GPT-4~\cite{gpt4}, have made remarkable strides in mathematical reasoning, especially in solving mathematical word problems (MWPs)~\cite{roy2018mapping}. These models excel in comprehending complex numerical contexts and multi-step reasoning, with GPT-4 achieving a 97\% accuracy rate on the GSM8K dataset~\cite{zhou2023solving}. However, current research predominantly focuses on problem-solving abilities, including the correctness of answers and the consistency of intermediate reasoning steps. In the EdTech sector~\cite{xu2024foundation,wang2021educational,wang2024large}, problems obtained from various sources often come with detailed solutions, although the quality of these solutions can vary. Therefore, the accuracy of answers is relatively straightforward to assess. Instead, identifying and correcting errors—a crucial aspect for enhancing educational efficiency—has been underestimated or overlooked. To maximize educational effectiveness, it is essential to accurately identify and address individual student problems. Given the rapid advancements in deep learning, we explore whether the potential of large models in mathematical reasoning can be harnessed to realize the following vision:

\begin{tcolorbox}[top=1pt, bottom=1pt, left=1pt, right=1pt]
\begin{center}
\textit{Can we leverage the mathematical reasoning power of advanced LLMs to create an AI-driven virtual teacher that autonomously analyzes student errors and provides targeted instruction?}   
\end{center}
\end{tcolorbox} 
This innovative approach has the potential to drastically lower educational costs while simultaneously improving teaching efficiency and accessibility. To achieve this, we present the design, implementation, and deployment of a virtual AI teacher focused on student error correction, as depicted in Figure~\ref{fig:function}. This system uniquely incorporates student drafts as the primary analysis target, deepening the understanding of each student's learning process. By harnessing the power of large language models and advanced prompt engineering, the system evaluates individual student performance, pinpointing and analyzing the root causes of errors with high accuracy. To optimize cost and efficiency, we maintain an error pool that catalogs historical errors, enabling the system to minimize computational demands when matching student answers to known errors. Furthermore, to enhance student engagement, we developed a real-time, multi-round AI dialogue system that allows students to effectively inquire about the knowledge points related to their problems. To the best of our knowledge, no other AI-based learning system offers such comprehensive feedback.

Compared to traditional teachers and machine learning-based error correction systems, our \abbr system offers the following advantages:

\begin{enumerate}
    \item \textbf{Significant Reduction in Educational Costs:} Currently, human teacher expenses constitute a large portion of course costs. By partially replacing human instruction with AI-based teaching, costs can be significantly reduced.
    \item \textbf{High Scalability:} Typically, only teachers with advanced knowledge in a field can provide excellent educational resources. However, the \abbr system, leveraging the knowledge capabilities of existing large models, can bypass this requirement and extend to various subjects and grade levels.
    \item \textbf{Generalizability}: Our \abbr system offers unparalleled capabilities in open-world error detection and recommendation, demonstrating superior generalization and practicality.
    \item \textbf{Flexible Educational Process:} The \abbr system allows students to start and stop their learning sessions at any time, ensuring that their educational experience remains uninterrupted by external factors.
\end{enumerate}

\noindent Finally, we summarize the contributions of this paper:
\begin{enumerate}
    \item To our knowledge, we are the first to introduce multimodal data with students' draft images, into error localization and analysis. Empirical evidence shows that using multimodal large models is significantly more advantageous than directly applying other AI technologies, such as using LLMs to analyze answers or applying OCR techniques to the draft first followed by LLM analysis.
    \item To build the \abbr system, we applied many advanced techniques, including complex prompt engineering, dual-stream models for error analysis, and internal knowledge point graph comparisons. This allows the system to interact conveniently with users and provides error localization, error analysis, problem feedback, and flexible dialogue functions.
    \item In terms of efficiency, we constructed an error pool to store historical answers and corresponding errors, avoiding repeated calls to large models for the same input, significantly improving access efficiency and reducing costs.
    \item The \abbr system has been deployed in Squirrel AI’s learning machines and has accumulated millions of usage samples. In this paper, we present partial evaluation results, where the error analysis accuracy of the \abbr system exceeds 75\%. Post-deployment, students' learning efficiency and mastery of knowledge points have noticeably increased. Additionally, a satisfaction survey of sales personnel showed an overall satisfaction rating of over 8 (out of 10), indicating a strong positive impact on product sales.
\end{enumerate}

\section{Related Work}

\textbf{LLM and MLLM for Mathematics.} In recent years, both LLMs~\cite{gpt4,brown2020language,touvron2023llama,jiang2023mistral,chiang2023vicuna} and MLLMs have experienced rapid development. With the emergence of models such as GPTs, Llama, and Vicuna, MLLMs have similarly undergone significant evolution~\citep{yin2023survey,zhang2024beyond,zhang2024mme,fu2024mme}. Mathematical ability has consistently been a specific challenge in evaluating these large models, with some efforts focusing on addressing mathematical problems through fine-tuning or large-scale training on relevant mathematical data\footnote{https://github.com/QwenLM/Qwen2-Math}. However, the majority of current work primarily emphasizes the ability to solve the problems correctly. For educators, determining the correct answers is often the simplest part, as we typically have ground truth answers before creating questions. The more challenging aspect is analyzing students' errors and providing appropriate feedback. 

\textbf{Error localization and analysis} is a problem that has been recognized for a long time. For example, some early systems were BUGGY~\cite{brown1978diagnostic}, which constructed an error database using a large amount of student error data and then used this database to predict the errors other students might make. Other methods have attempted to automatically generate error-detection rules using machine learning, and then perform error analysis based on these rules. However, these methods have a significant drawback: they struggle to handle new errors that are not captured by the existing databases or rules. Even though recent work has expanded the scope of analysis through complex module combinations~\cite{feldman2018automatic}, they are still limited to monitoring a narrow range of error categories. Intuitively, the powerful few-shot learning capabilities of multimodal large models, combined with their extensive internal knowledge, hold promise for achieving open-world error identification and analysis. However, as far as we know, currently there is no work dedicated to using large models specifically for error localization and analysis. A recent benchmark claims to be designed for evaluating this function~\cite{li2024evaluating}, but its problems are relatively simple and artificially constructed from public datasets, which are still far from the reality of actual error analysis.

We attempt to build the \abbr system using various engineering tools, leveraging the cognitive capabilities of large models to create an AI tool that can autonomously complete analysis and provide recommendations. To our knowledge, this is the first educational system capable of open-world, autonomous analysis and inference. After practical deployment and extensive user experience evaluations, \abbr significantly improved student learning efficiency and received widespread recognition from sales personnel.

\section{Use of AI Technology}
\begin{figure*}
    \centering
    \includegraphics[width=0.9\linewidth]{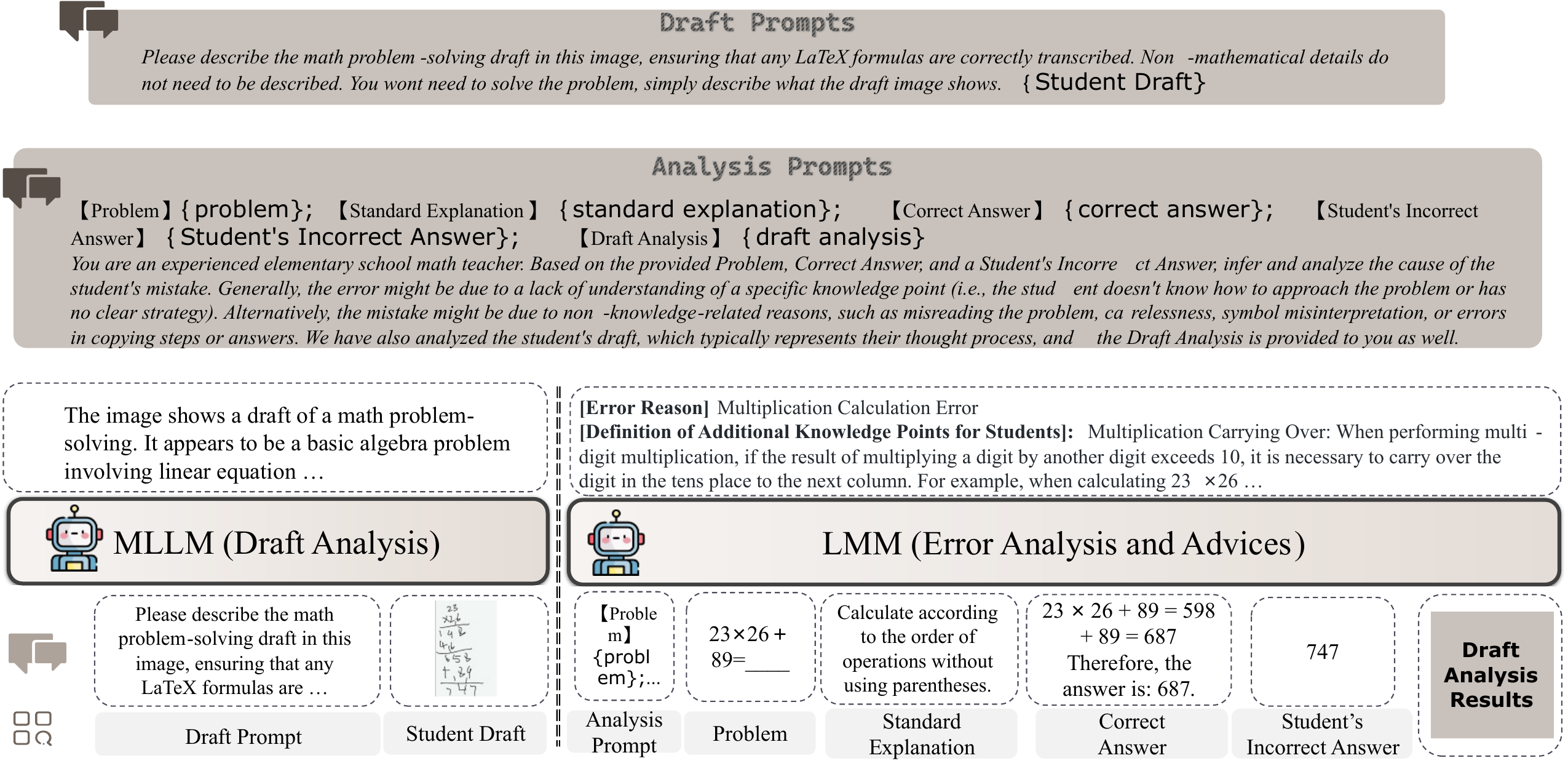}
    \caption{\textbf{Architecture of \abbr}. The \abbr framework is a multi-agent system that begins by analyzing a user-inputted draft image using our specialized Draft Prompt mechanism. This draft analysis, alongside student responses, problem explanations, and correct answers, is then formatted according to our predefined structure and inputted into another language model. The system ultimately produces a detailed error analysis and recommendations, guiding students on how to correctly approach and solve similar problems.}
    \label{fig:framework}
\end{figure*}

\subsection{Multi-modal Data Collection}

Using large models to judge errors in mathematical problem-solving steps is a very intuitive idea. In the initial phase of our project, we also attempted to use LLMs to directly analyze incorrect answers, including various complex prompt engineering techniques, such as inputting knowledge points, answers, explanations, and student answers. However, LLMs consistently failed to provide reliable error cause analysis and could not be used effectively. Due to the limited information available, LLMs could only guess where the student's intermediate steps went wrong based on the final answer, leading to a significant margin of error in these guesses. Therefore, we began advocating for students to fully document their calculation steps and upload their draft images to the backend, rather than just submitting the incorrect answers. As shown in Figure~\ref{fig:function}, when students do not upload their intermediate steps on the draft, we request that they redo the problem. Since implementing the policy of requiring students to write drafts, the proportion of students submitting detailed problem-solving processes has increased from 5\% four months ago to approximately 60\%. Enforcing the standard of submitting drafts not only improves students' focus while solving problems but also provides us with more options for subsequent processing. To date, we have collected over 24 million student drafts, a choice of immense value to the education industry. In our project, we have also verified that analyzing drafts containing intermediate steps greatly enhances the effectiveness and reliability of applying large models for error analysis.

\subsection{Error Cause Analysis and Understanding}

\subsubsection{Dual-Stream Large Model Error Cause Analysis}
The framework is illustrated in Figure~\ref{fig:framework}, which includes two models that perform draft analysis and error cause analysis with recommendations. When a student's answer is marked incorrect, their draft, the problem, the problem explanation, and the correct answer are processed step by step, resulting in an analysis of the error causes and suggestions for follow-up learning.

\textit{1. Effective Utilization of Draft Data:} We explored various approaches to effectively use draft data. Initially, before multimodal large models had achieved their current performance, we attempted to use existing OCR tools to analyze the drafts and then input them into an LLM for analysis. However, this approach was significantly limited by the performance of OCR tools, which could only recognize text and numeric information. These tools struggled to effectively interpret the spatial structure of the student's draft data and mathematical symbols, resulting in poor outcomes. Given the gradual improvement in the image understanding capabilities of multimodal large models, we later integrated these models into our system. As shown in Figure~\ref{fig:framework}, we designed specific prompts for draft analysis and input them, along with the student drafts, into the multimodal large models. These models were then able to provide us with specific information contained in the drafts as well as descriptions of the problem-solving process.

\textit{2. Generating Detailed Error Causes and Suggestions:} With the initial draft analysis complete, we naturally input the draft summary directly into the LLM, enabling it to infer the student's error causes based on the intermediate steps. However, at this stage, the information provided by only using the draft data and comparing the incorrect/correct answers was still insufficient for the LLM to perform a comprehensive error cause analysis. To overcome this limitation, we systematically experimented with nearly all entries in our database and identified the optimal system prompt. As illustrated in Figure~\ref{fig:framework}, we combined the problem, its solution, the correct answer, the answer explaination, the student’s incorrect answer, and the draft analysis according to the structure shown in the figure. We also added carefully designed guiding words, allowing the LLM to have a comprehensive understanding of both the correct answer and the student's complete steps. This approach enabled the LLM to generate a detailed analysis of the error causes and provide suggestions for the student.
\begin{figure}
    \centering
    \includegraphics[width=\linewidth]{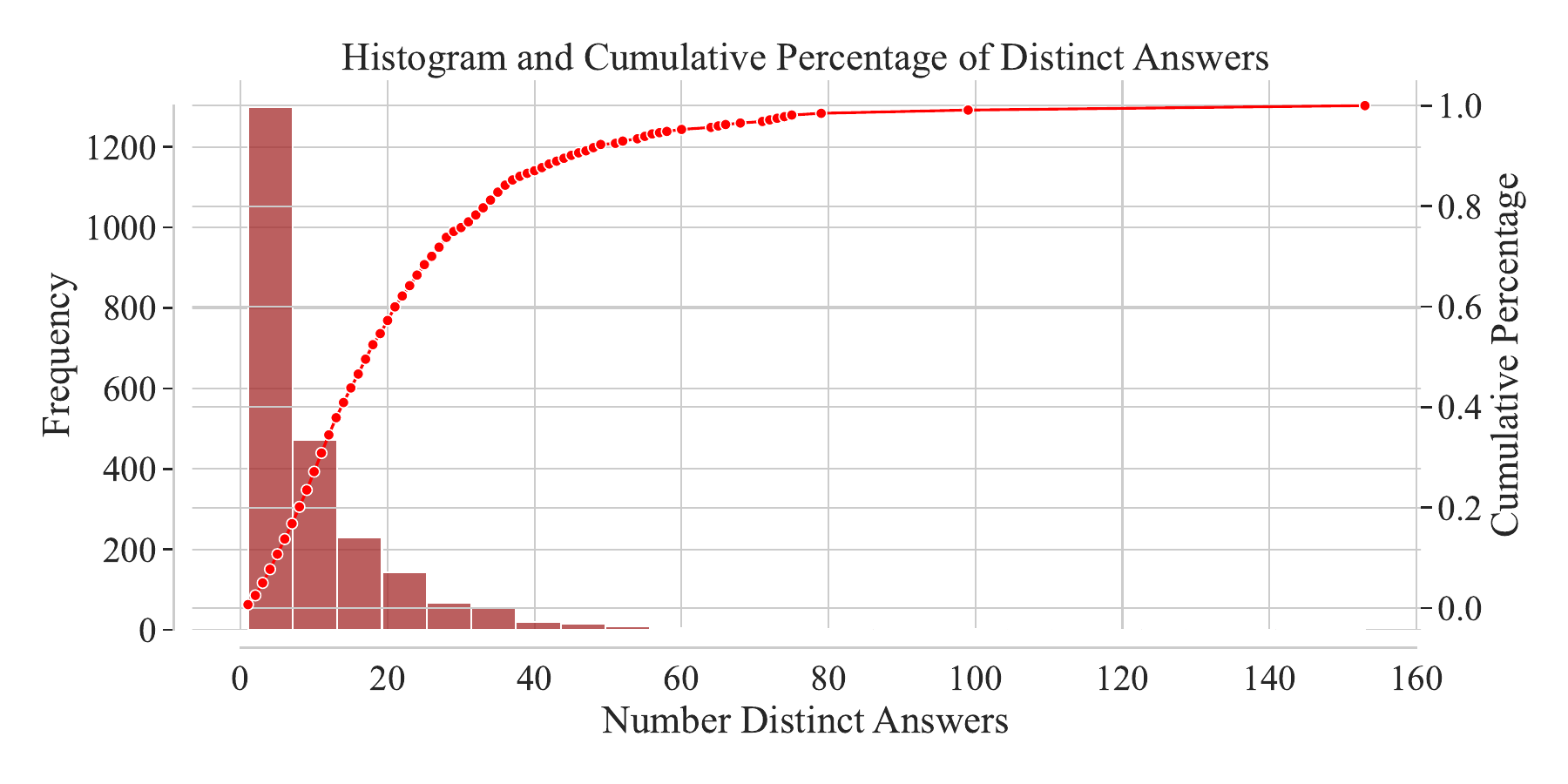}
    \caption{Distribution of the number of different incorrect answers across 2,326 questions.}
    \label{fig:ans_dis}
\end{figure}
\subsubsection{Error Cause Analysis Based on the Error Pool}
Using dual-stream large models to analyze each student's answer can yield good results, but given our large user base, there could be scenarios where tens of thousands of students simultaneously send requests. In such cases, calling the LLM becomes a performance bottleneck, leading to significant efficiency issues. Additionally, we observed that the distribution of student errors follows a long-tail pattern. We analyzed over 2,000 representative errors from the error pools of the top 800 students ranked by error rate between April and May 2024. As shown in Figure~\ref{fig:ans_dis}, most incorrect answers fall into fewer than 40 categories, and the error causes for each incorrect answer is generally the same. For example, in Figure~\ref{fig:framework}, the only reason for calculating \(23 \times 26 + 89\) as 598 is forgetting to add the final 89. For such errors, it is unnecessary to call the large model for re-analysis. Instead, we can use the analysis and responses stored in the historical error pool to provide suggestions and feedback to the student. This approach not only avoids the variance introduced by multiple calls but also improves the model's efficiency. Specifically, our error cause analysis using the error pool involves two steps:

\begin{figure}
    \centering
    \includegraphics[width=\linewidth]{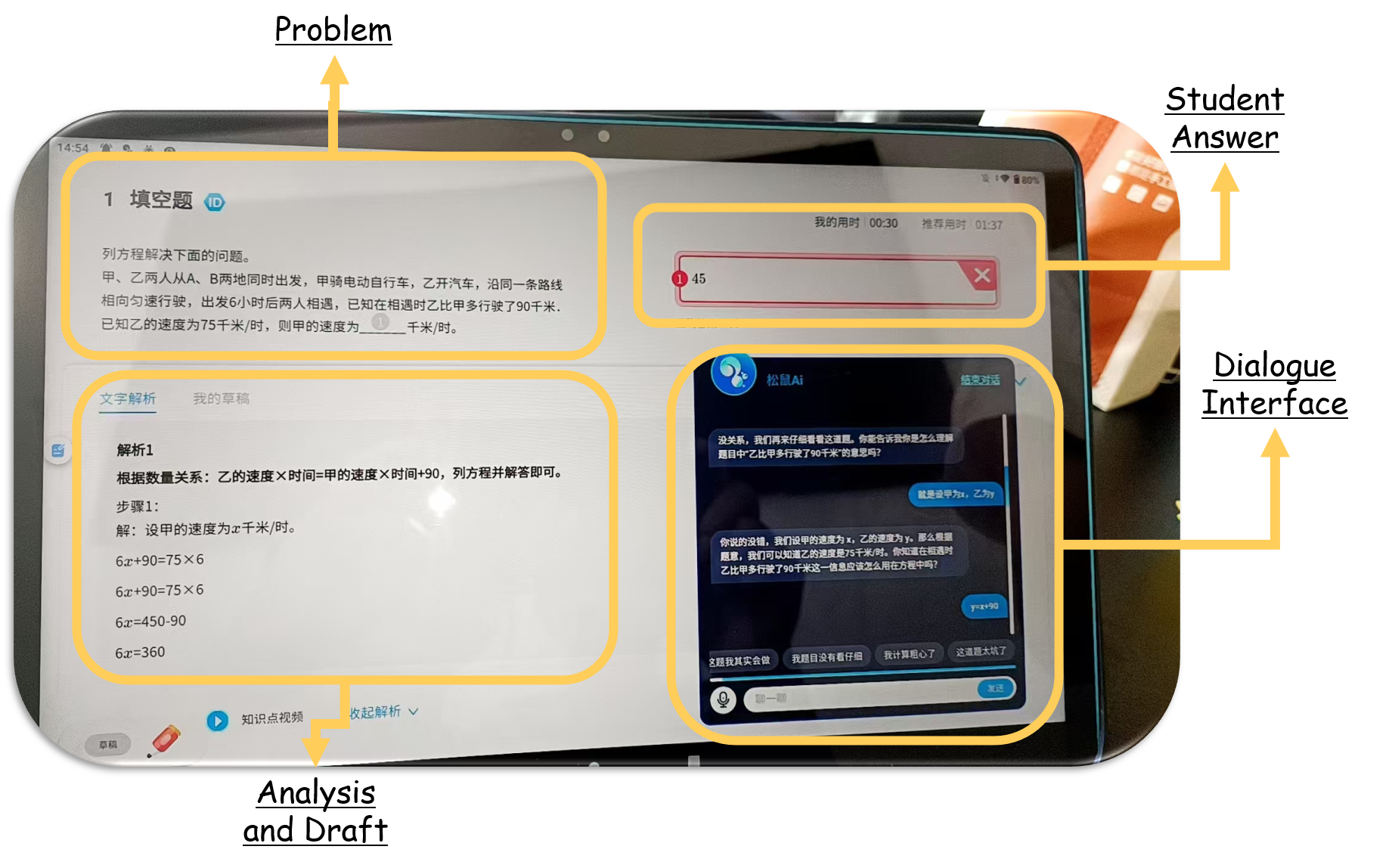}
\caption{\textbf{The \abbr system deployed in the Squirrel Ai Learning machine interface}, displaying the problem statement, solution explanation, student's draft, and AI dialogue options.}    \label{fig:squirrel_ui}
\end{figure}

1) \textit{Error Pool Matching:} Our error pool stores pairs of question IDs and student answers as hash keys, with each pair having a unique hash key. This means that we assume for each question, if students provide the same incorrect answer, their intermediate steps are likely similar, leading to the same error cause. This rule-based matching approach is common in traditional error analysis tools~\cite{brown1978diagnostic}. The assumption is reasonable and has been validated with an expert for lower-grade cases (e.g., K5). For more complex cases, the expert suggested that we might introduce additional hash keys, such as intermediate answers, if feasible. Nevertheless, the assumption remains valid for a substantial portion of problems. As shown in Figure~\ref{fig:function}, when a student provides an incorrect answer, if the question ID and student answer match an entry in the error pool, the system directly returns the precomputed error cause. Otherwise, the answer is passed on to our dual-stream large model for analysis.

\textit{2) Error Pool Update:} Updating the error pool can be straightforward: whenever a new question ID and student answer are encountered, the error pool is expanded. However, sometimes students provide completely random answers that are of no value. Additionally, for more complex problems, there can be many possible errors. Without restrictions, this could lead to rapid expansion of the error pool, reducing retrieval efficiency. Therefore, we have implemented the following constraints for updating the error pool:

\begin{enumerate}
    \item \textbf{Quality:} It is challenging to rely solely on answers. Fortunately, since we require students to upload draft content, we first ask the multimodal model to score the draft based on clarity, spatial utilization, organization, consistency, correction traces, and neatness. Drafts with low scores generally reflect random scribbles by students and provide little valuable reference. We only update the error pool with new question ID-student answer pairs when the draft quality meets our standards.
    \item \textbf{Quantity:} We limit the error pool for each problem to a maximum of 100 entries to reduce the retrieval burden.
\end{enumerate}

The primary benefit of the error pool is the reduction in computational costs. Let \(N\) be the total number of problems in our system and \(K\) be the maximum number of distinct student answers per problem. The total number of LLM API calls (as well as the amount of tokens consumed) is then upper-bounded by \(O(NK)\), independent of the number of users, trigger frequency, etc. This is especially advantageous when budgeting for multimodal LLM API calls, which are typically more expensive than pure text calls.

\begin{figure*}[!ht]
\centering
\begin{minipage}[t]{0.22\linewidth}
\raggedleft 
\includegraphics[width=\linewidth]{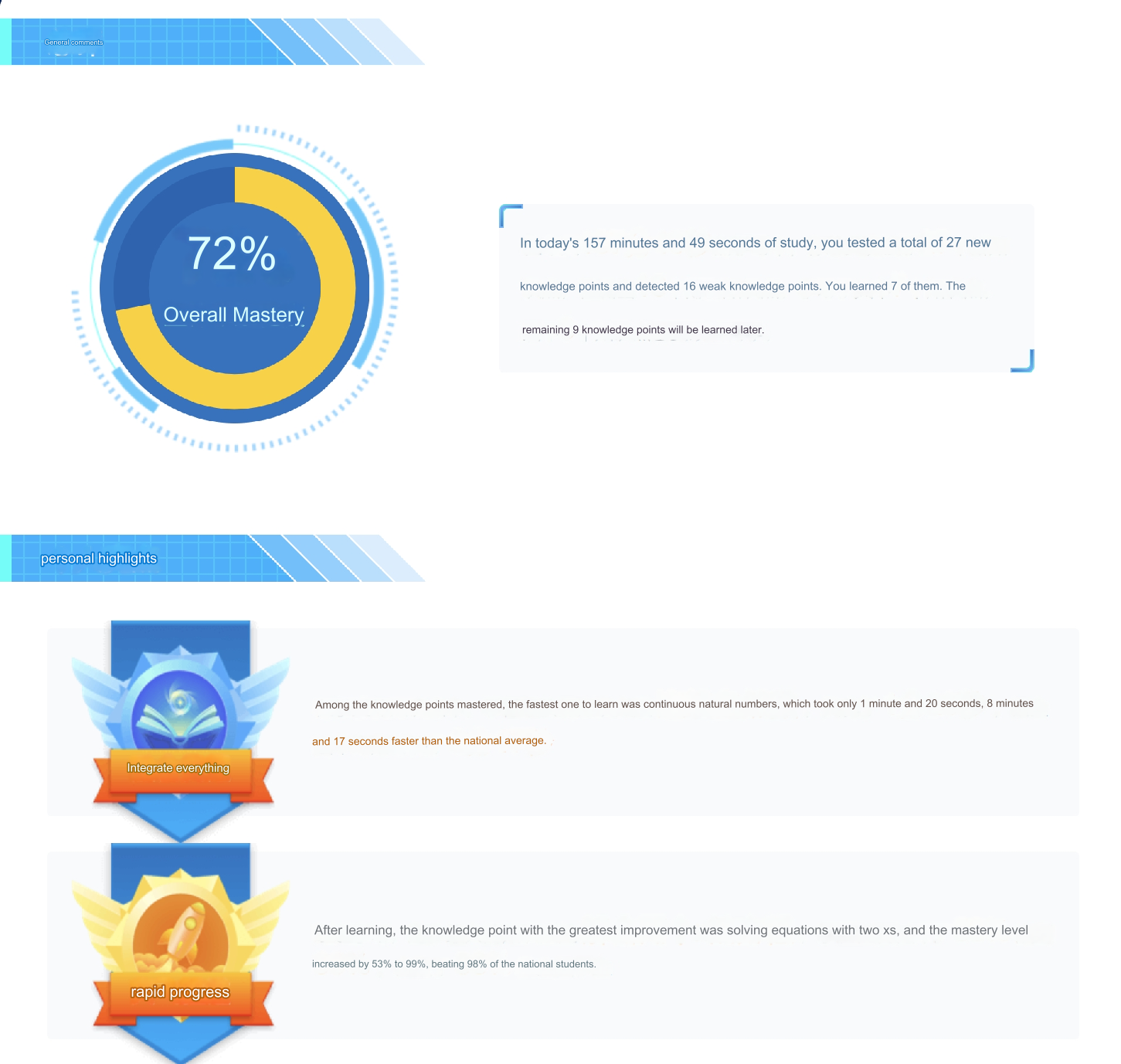}
\end{minipage}%
\begin{minipage}[t]{0.27\linewidth}
\centering
\includegraphics[width=\linewidth]{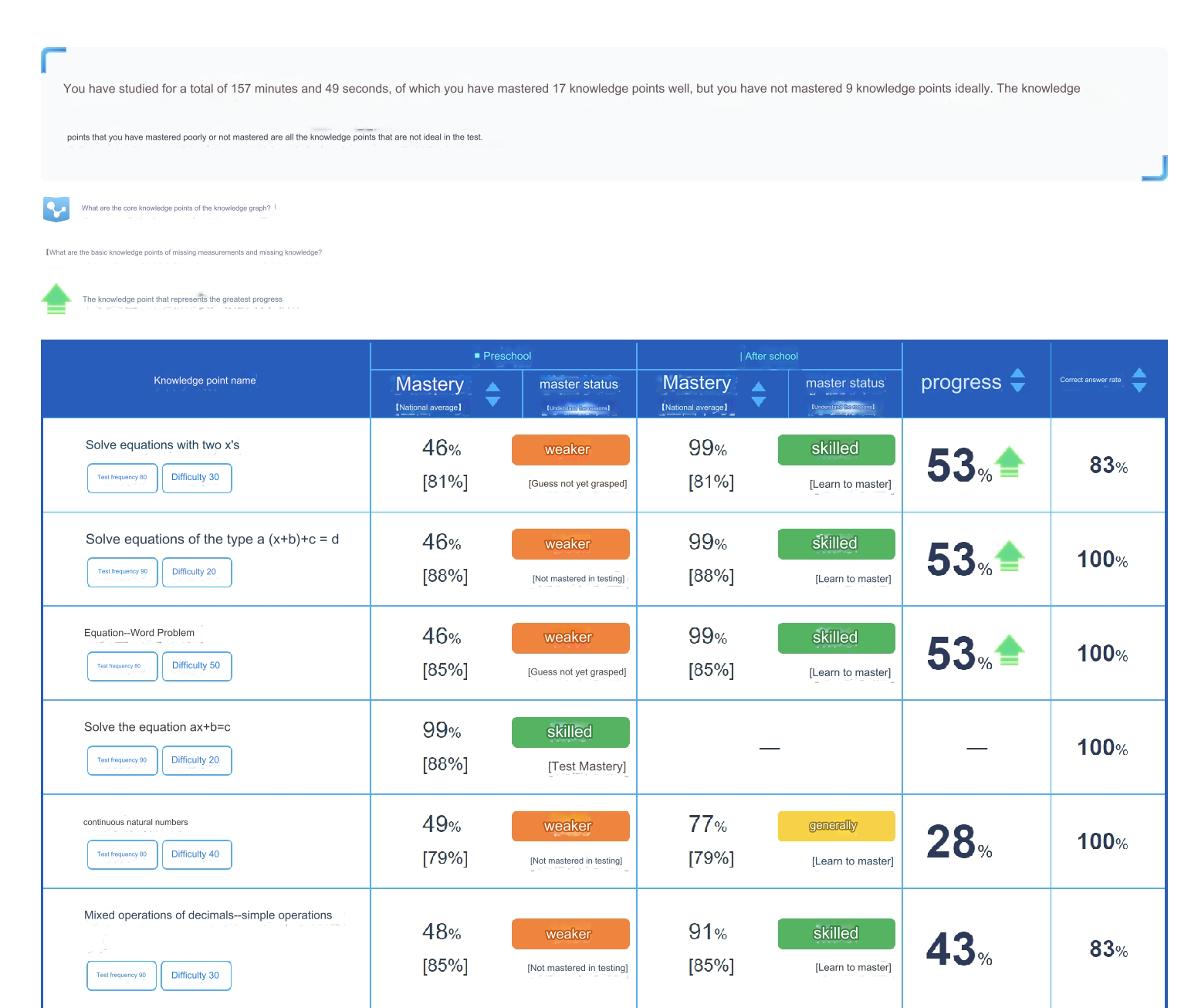}
\end{minipage}%
\begin{minipage}[t]{0.39\linewidth}
\raggedright
\includegraphics[width=\linewidth]{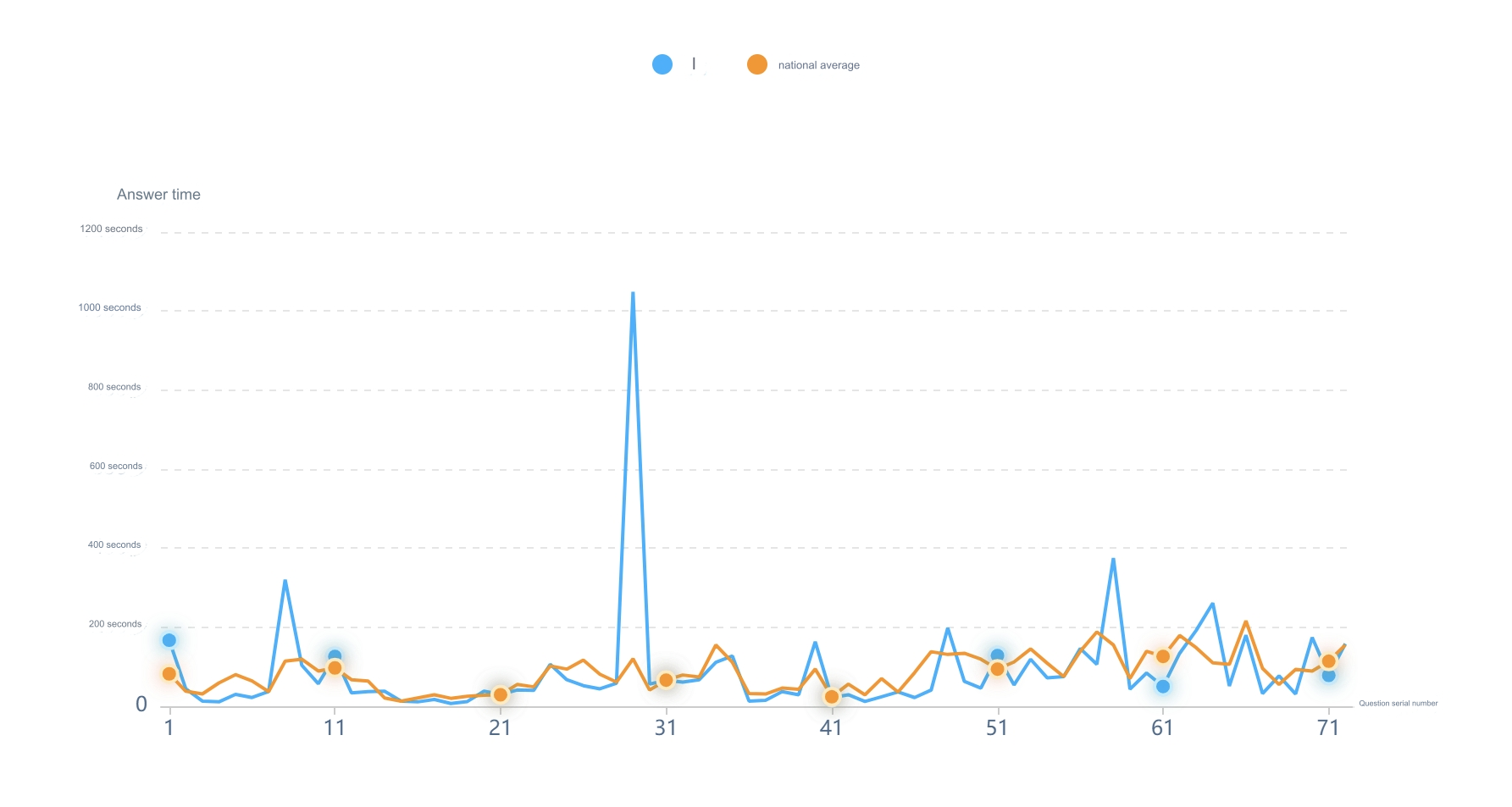}
\end{minipage}%
\centering
\caption{\textbf{\abbr Learning Summary Pages:} (left): Overview of the student's learning content for the day, including study duration and associated knowledge points. (middle) Detailed analysis of the student's mastery across various relevant knowledge points. (right) Comparative analysis of the student's problem-solving time versus the national average.}
\label{fig:ui_summary}
\vspace{-0.1cm}
\end{figure*}
 
\section{Dialogue System Design}
To date, we have developed a robust and efficient error analysis and recommendation tool. However, how to interact effectively with users remains a critical challenge. In the process of educating students, it is not advisable to directly provide the complete error analysis and answers. Instead, it is more beneficial to guide students step by step to discover the issues themselves, thereby enhancing the learning effect. To achieve this, we have designed specific prompts for the large model. Similar to the initial error analysis stage, we input the problem, explanation, student’s answer, and the model’s error analysis and suggestions into the LLM. However, unlike the former, we impose a series of requirements on the LLM in this stage: 1) Avoid directly providing the answers; instead, guide students to find the answers themselves through questions and hints. 2) Adhere to a guided and heuristic teaching approach, encouraging students to engage in self-directed learning and critical thinking. 3) Maintain relevance in the dialogue, focusing on the issues and challenges the student is currently facing. 4) Respect the student's learning pace, avoiding overly rapid progression. 5) Refrain from discussing topics unrelated to learning.

With these requirements in place, our dialogue system can provide guided instruction to students, building upon the foundation of error analysis.

\section{Deployment and Application}

We illustrate the student interface post-\abbr deployment in Figure~\ref{fig:squirrel_ui}, highlighting the key components such as the dialogue box, problem and solution explanation, draft, and student's answer. After each learning session, we can track the student's study time, mastery of each knowledge point, and compare their time usage with that of students nationwide (Figure~\ref{fig:ui_summary}). Note that while the deployed environment predominantly features Chinese content, we have translated the interface into English for easier review in Figure~\ref{fig:ui_summary}.

The \abbr system is flexible with different backbone large models and highly efficient in the deployed products. When it is powered by GPT-4o, the average number of input tokens is 1865, the average number of output tokens is 81, and the average response time is 1955 ms.
As of now, the \abbr system has achieved widespread deployment, covering 35 provinces and 338 cities, significantly impacting the analysis and guidance of elementary school students' mathematical errors. With over 100,000 usage records, this system has become an integral tool for educators, demonstrating its extensive reach and effectiveness in addressing student learning challenges across a vast geographical area.


\begin{figure*}[!htp]
    \centering
    \includegraphics[width=0.9\linewidth]{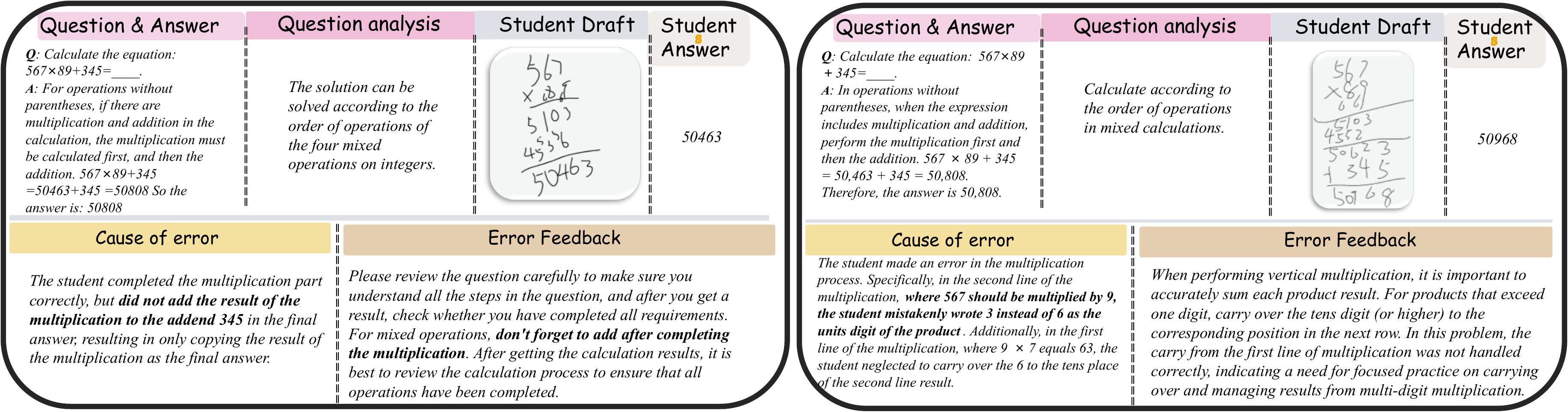}
    \caption{\textbf{Examples of Error Cause Analysis.} (Left) A successful case where the \abbr system correctly identified that the student forgot to add 345, resulting in the suggestion "don't forget to add after completing the multiplication." (Right) An unsuccessful case where the model incorrectly identified the error, leading to an inaccurate analysis of the multiplication step.}
    \label{fig:examples}
\end{figure*}

\section{Experiments}

\subsection{Human Evaluation of Error Analysis Capabilities}\label{sec:human_eval_error}

To validate the effectiveness of our \abbr system's error analysis capabilities, we conducted an experimental analysis on a selection of representative and commonly mistaken question types and categories from over $200K$ usage samples. The data selection logic was as follows: (1) Questions that students repeatedly answered incorrectly, meaning the question was re-recommended to the student, and they continued to make errors; (2) Questions with a relatively high average error rate, specifically those ranking high in overall incorrect response rates. We applied a weighted ranking based on criteria (1) and (2) to select the top 420 questions. 

These selected sample questions, along with the error analysis and recommendations generated by \abbr, were then evaluated by experts who rated them on a binary scale, with 1 
indicating success and 0 indicating suboptimal performance. Ultimately, $78.3\%$ of the samples passed expert review, and the generated recommendations were also well-received by the experts. Although 78.3\% may not appear to completely resolve issues related to error cause analysis, it's important to note that our test data had been cleaned, meaning we selected the most challenging problems for evaluation. Moreover, the remaining 21.7\% of the samples still provided some reference value, even if they did not meet the accuracy standards set by expert review. This demonstrates that our system is already suitable for daily use and has achieved notable success on the Squirrel AI platform.

In Figure~\ref{fig:examples}, we present one example each of a successful and a failed case. We observed that inaccuracies in error cause analysis were sometimes due to the multimodal large model misinterpreting the draft, while in other instances, the LLM struggled to summarize or infer the fundamental reason behind the error. Therefore, improvements in both models could significantly enhance the quality of our \abbr system, indicating considerable potential for further development.

\subsection{The Impact of \abbr on Student Learning Outcomes}

\begin{table}[!t]
\caption{Metric Abbreviations and Definitions}
\centering
\resizebox{\linewidth}{!}{%
\begin{tabular}{l|p{9cm}}
\hline
\textbf{Abbreviation} & \textbf{Definition} \\ \hline
NIACT & The cumulative number of incorrect answers in the knowledge point associated with the current question during this learning session. \\ \hline
NQCT & The cumulative number of questions in the knowledge point associated with the current question during this learning session. \\ \hline
ARCT & The average correct answer rate for the knowledge point associated with the current question during this learning session. \\ \hline
NVRS & The number of times the student has rewatched videos and relearned the knowledge point associated with the current question due to low mastery during this learning session. \\ \hline
\end{tabular}
}
\label{table:metrics}
\end{table}
To validate the impact of \abbr on students' abilities, we conducted a statistical analysis of learning data from 5,600 students in Squirrel AI, categorized by different metrics. We classified the data based on whether students used our \abbr, and whether their subsequent attempts on the same question were correct after interacting with the system. Specific metrics are detailed in Table \ref{table:metrics}, and the results are presented in Table \ref{tab:main_results}.

Overall, after using our system, even if students did not immediately learn the correct answer (not effective), their error rate on related knowledge points decreased by 15\%. This indicates that our prompts and suggestions significantly enhance students' understanding of the relevant concepts. Additionally, students' learning efficiency (NQCT) and problem-solving accuracy (ARCT) showed noticeable improvement. When students engaged in highly effective communication with the system, the error rate on relevant knowledge points dropped by 37\% compared to those who did not use our system, with further improvements in learning efficiency and accuracy. Moreover, the NVRS, which measures the number of times students needed to rewatch videos to grasp related knowledge, decreased by 61.5\%.


\begin{table}[] 
\caption{\textbf{Impact of \abbr on Learning Outcomes.} The "Conversation" metric indicates whether students used our system to correct their mistakes, while "Effective" reflects the effectiveness of student-system interactions—specifically, whether students were able to produce correct answers after discussing with the system.}\label{tab:main_results}
\resizebox{\linewidth}{!}{%
\begin{tabular}{cccccc}
\hline
\textbf{Conversation} & \textbf{Effective} & \textbf{NIACT} & \textbf{NQCT} & \textbf{ARCT} & \textbf{NVRS} \\ \hline
× & × & 4.96 & 9.99 & 0.52 & 0.39 \\
\checkmark  & × & 4.18 & 9.59 & 0.59 & 0.30 \\
\checkmark  & \checkmark  & 3.12 & 7.39 & 0.62 & 0.15 \\ \hline
\end{tabular}%
}
\end{table}


\textbf{The Impact of Dialogue Quality on Learning Outcomes} In practical applications, we observed a significant amount of ineffective dialogue. Some conversations were very brief, consisting of only a few characters and lacking clarity, while others contained excessive characters, often resulting from random input by students. We defined moderate dialogue as having a total character count between 15 and 120. Moderate dialogue accounted for only 61.67\% of all dialogue content. Table~\ref{tab:length_diag}  shows the impact of three types of conversation on the number of times students need to engage in repeated learning of a single knowledge point. In our educational system, each knowledge point is associated with various videos and exercises. Learning only ceases when students have repeatedly studied a knowledge point and their accuracy in solving problems reaches a certain level. The more times students need to repeat their learning, the lower their learning efficiency. As shown, even a small amount of dialogue leads to improved learning efficiency, while moderate interaction maximizes it. Beyond this point, additional dialogue contributes little to further improvements. In the optimal scenario, learning efficiency can be increased by up to 40.5\%.

\begin{table}[!t]
\caption{Impact of the \abbr Platform on Foundational Learning Sessions. The value in "average" refers to the number of times students need to engage in repeated learning of a single knowledge point. In our educational system, the larger the value, the lower the learning efficiency.}
\label{tab:length_diag}
\centering
\begin{tabular}{cccc}
\hline
\textbf{Conversation} & \textbf{Effective} & \begin{tabular}[c]{@{}c@{}}\textbf{Conversation} \\  \textbf{Quality}\end{tabular} & \textbf{Average} \\\hline 
× & × & No Dialogue & 9.81  \\
\checkmark  & × & Too short & 7.19 \\
\checkmark  & \checkmark  & Too short & 6.23 \\
\checkmark  & × & Moderate & 6.54 \\
\checkmark  & \checkmark  & Moderate & 5.85 \\
\checkmark  & × & Too long & 6.46 \\
\checkmark  & \checkmark  & Too long & 5.83 \\ \hline
\end{tabular}%
\end{table}

\begin{table}[!t]
\caption{Impact on Win Rate by Ablating Draft, Problem Content, Correct Solution, and Student Answer. The win rate is calculated by comparing outputs when all elements are included versus when one element is ablated. A higher win rate indicates that excluding the element has a greater impact on performance. }
\centering
\begin{tabular}{ccccc}
\hline
\textbf{w/o}     & \textbf{Draft} & \textbf{Problem} & \textbf{Solution} & \textbf{Answer} \\ \hline
Win Rate       & 0.61  &     0.67              &   0.77       & 0.70           \\ \hline
\end{tabular}
\label{tab:ablation}
\end{table}

\subsection{Ablation Study}
To assess the impact of various elements on overall performance, we conducted an ablation study by systematically removing one element from the prompt at a time—Draft, Problem Content, Correct Solution, or Student Answer. We then measured the resulting win rate. The win rate is defined as the proportion of cases where the output with all elements included outperforms the output with one element removed. A senior education expert conducted the comparison, ensuring a rigorous evaluation of the results. The study included 418 records, and the reported win rates represent the average across these cases.

We present the result in Table \ref{tab:ablation}. Student drafts offer direct insight into their thought processes, enabling the LLM to trace the causes of errors. Without these drafts, the LLM's ability to accurately diagnose the cause of errors is significantly hindered. The problem content provides the necessary context that frames the LLM's reasoning, allowing it to interpret the task correctly. The correct solution, on the other hand, serves as a critical reference point for identifying deviations in student responses. Finally, students' wrong answers are pivotal for error analysis, as they reveal specific misconceptions or gaps in understanding that need to be addressed. Each element plays a unique and indispensable role in the LLM's comprehensive error analysis.

\begin{figure}
    \centering
    \includegraphics[width=1.0\linewidth]{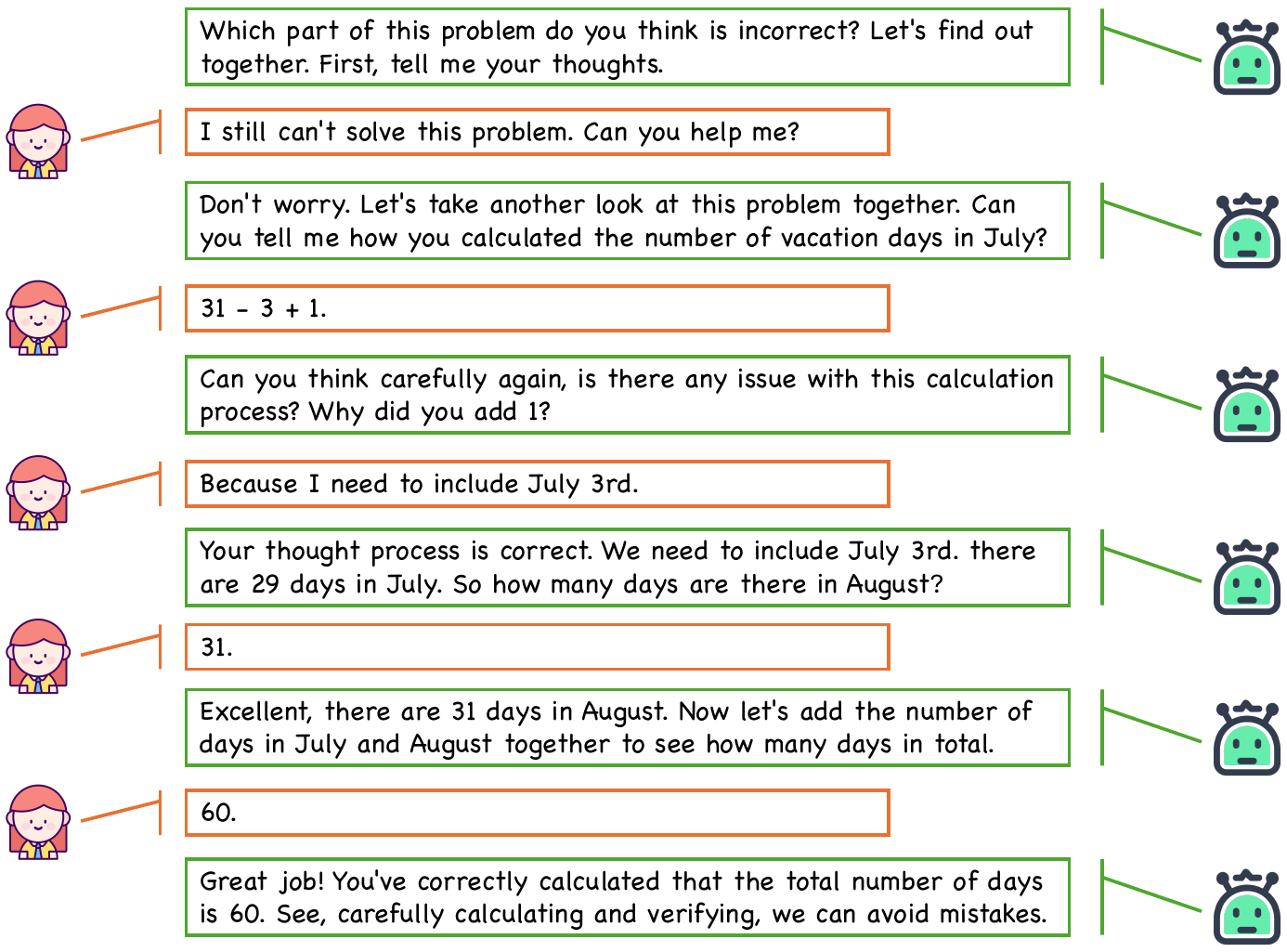}
    \caption{Error analysis dialogue with a real student. }
    \label{fig:case_study}
\end{figure}


\subsection{Case Study}
In Figure \ref{fig:case_study}, we present a real-world example of an error analysis dialogue. Before the dialogue, The agent conducted the error analysis as outlined in Figure \ref{fig:framework} and successfully identified that the error was caused by a calculation oversight. The dialogue begins with an open-ended question posed by the agent, designed to encourage the student’s independent problem-solving and critical thinking. After that, the agent leads the discussion by methodically solving the problems step by step. With the agent's assistance, the student learns to break down the problem of ``\emph{calculating days between dates}'' into calculating the days within two separate months and finally derive the correct answer. At the conclusion of the dialogue, the agent encourages the student to be more attentive in future calculations, highlighting that the error was due to a simple oversight. This case shows that the error analysis dialogue provides a more user-friendly protocol between the system and students, facilitating easier learning from mistakes.

\subsection{Survey on Sales Personnel Satisfaction}
\begin{figure}
    \centering
    \includegraphics[width=\linewidth]{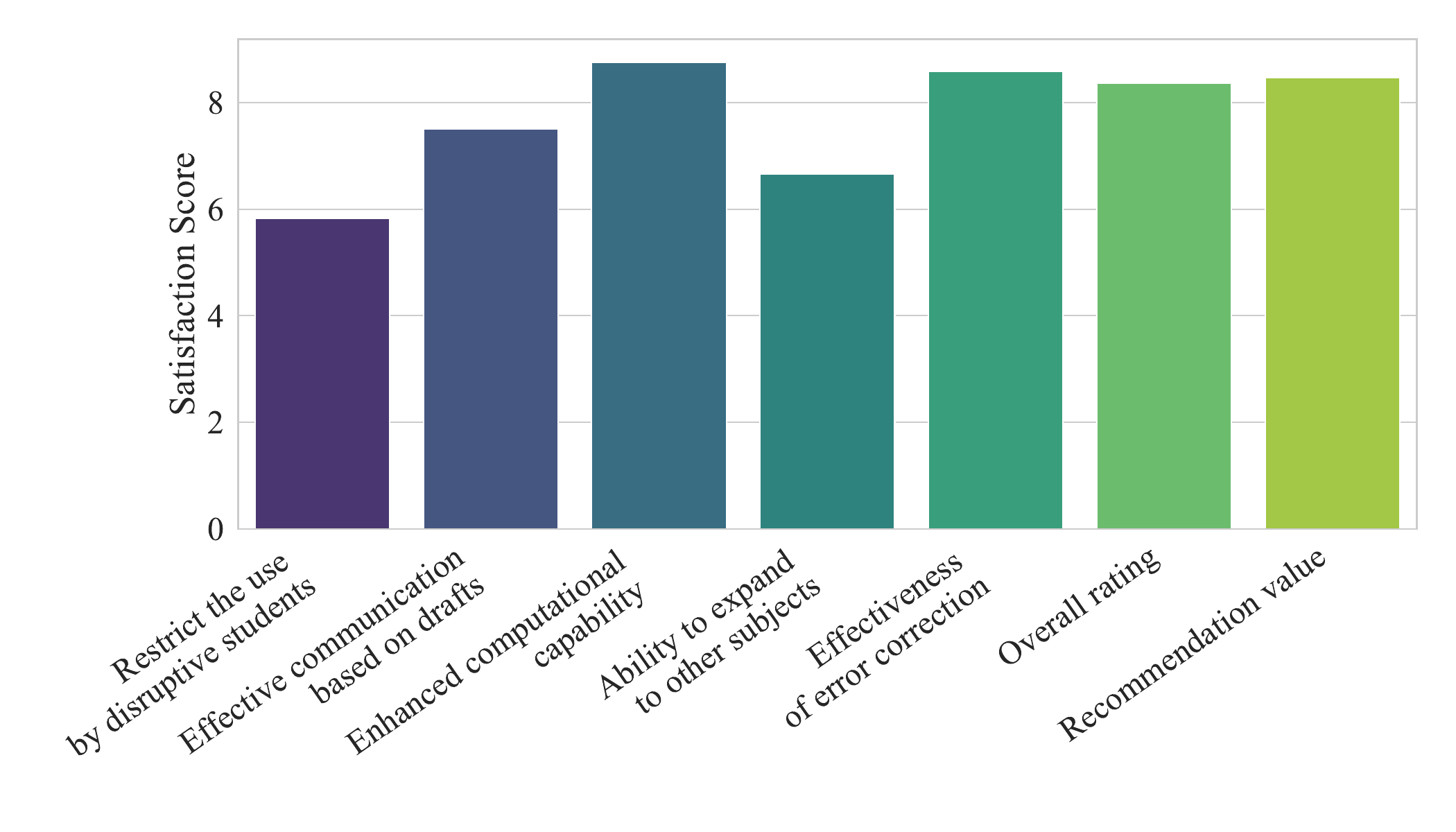}
    \caption{\textbf{Results of the satisfaction survey among different personnel types at Squirrel AI}, showing high effectiveness in preventing uncivil learning behaviors, strong scalability across disciplines and grade levels, and overall high ratings in error correction and system performance.}
    \label{fig:satis}
\end{figure}

To gain a better understanding of user experiences, we conducted a satisfaction survey among four types of personnel at Squirrel AI: store managers (14.71\%), dealers (23.53\%), supervisors (52.94\%), and salespeople (8.82\%)\footnote{All four categories of personnel possess a certain level of educational experience, enabling them to assess whether the deployed AI system genuinely holds educational value, while also providing feedback on user experiences through sales performance.}. The results in Fig.~\ref{fig:satis} show that nearly 60\% of respondents believe our virtual AI teacher effectively prevents uncivil learning behaviors (e.g., students inputting gibberish or engaging in personal attacks) and facilitates effective communication based on drafts. Over 66\% of participants acknowledged the scalability of the AI across different disciplines and grade levels. The effectiveness of the system's error correction was rated at 8.36 out of 10, with an overall score of 8.58. Sales personnel gave a recommendation score of 8.47 out of 10, highlighting the significant impact of Squirrel AI on educational efficiency and product upgrades.

\section{Conclusion}

In conclusion, the proposed Virtual AI Teacher System (VATE) represents a significant advancement in educational technology by effectively addressing the limitations of traditional error correction methods. By integrating multimodal data such as student drafts and utilizing dual-stream large language models, VATE not only enhances the accuracy of error cause analysis but also provides targeted instructional guidance. This approach allows for a scalable, cost-effective, and flexible educational process that can be generalized across various subjects and grade levels. The implementation of an error pool further optimizes the system's efficiency, reducing computational demands while maintaining high accuracy in error detection and feedback. Empirical results demonstrate that the VATE system significantly improves student learning outcomes, as evidenced by increased mastery of knowledge points and high satisfaction ratings. The success of VATE underscores the potential of AI-driven solutions in revolutionizing education, offering a model that can be adapted and expanded to meet diverse educational needs.

\section{Future work}
Building on the foundation established by error analysis, our future work will explore the development of a learning content recommendation system that leverages these insights to create more advanced adaptive learning systems~\cite{kdd24eduworkshop,li2024bringing}. Error analysis enables recommendation systems to become more targeted and goal-oriented, allowing for tailored interventions. For example, the system could suggest specific exercises to address non-knowledge-related errors, such as calculation mistakes or misunderstandings, while generating personalized learning paths for knowledge-based errors to enhance mastery of specific concepts. We plan to present this next step in our research in a follow-up paper, focusing on how integrating error analysis into recommendation systems can optimize learning outcomes and further improve student engagement and success.

\bigskip

\bibliography{aaai25}

\end{document}